\documentclass[conference,a4paper]{APSIPA2017}
\usepackage{multirow}
\usepackage[dvips]{graphicx}
\usepackage{amsmath}
\usepackage[psamsfonts]{amssymb}
\usepackage{amsxtra}
\usepackage{epstopdf}
\usepackage{threeparttable}

\begin{document}

\title{Enhanced Neural Machine Translation by Learning from Draft}

\author{%
\authorblockN{%
Aodong Li\authorrefmark{2},
Shiyue Zhang\authorrefmark{2}\authorrefmark{3} 
Dong Wang\authorrefmark{2}$^*$ and
Thomas Fang Zheng\authorrefmark{2}
}

\authorblockA{%
\authorrefmark{2}
Tsinghua University, Beijing, China 
}
\authorblockA{%
\authorrefmark{3}
Beijing University of Posts and Telecommunications, Beijing, China
}
$^*$Corresponding Author: wangdong99@mails.tsinghua.edu.cn
}
\maketitle
\thispagestyle{empty}

\begin{abstract}
Neural machine translation (NMT) has recently achieved impressive results. A potential problem of the existing NMT algorithm, however, is that the decoding is conducted from left to right, without considering the right context. 
This paper proposes an two-stage approach to solve the problem. In the first stage, a conventional attention-based NMT system is used to produce a draft translation, and in the second stage, a novel double-attention NMT system is used to refine the translation, by looking at the original input as well as the draft translation. This drafting-and-refinement can obtain the right-context information from the draft, hence producing more consistent translations. We evaluated this approach using two Chinese-English translation tasks, one with 44k pairs and 1M pairs respectively. The experiments showed that our approach achieved positive improvements over the conventional NMT system: the improvements are 2.4 and 0.9 BLEU points on the small-scale and large-scale tasks, respectively.
\end{abstract}

\section{Introduction}

Neural machine translation (NMT)\cite{1}\cite{3}\cite{4} has continuously gained attention from both academia and industry, and has obtained the state-of-the-art performance on many translation tasks, e.g., English-to-French, English-to-German, Turkish-to-English, and Chinese-to-English\cite{4}\cite{6}\cite{2}\cite{5}\cite{7}.

The basic NMT model employs a sequence-to-sequence architecture\cite{4}, where the meaning and intention of the source sentence is
encoded into a representation vector with fixed dimensions, by which the translation (target sentence) is produced word by word. 
This architecture was later extended to an attention-based model\cite{2}\cite{5}, which allows the decoder being aware of the location
that it should focus on at each decoding step. In a typical implementation of the attention-based NMT architecture, the encoder and decoder are both recurrent neural networks (RNN), where the hidden units are often some kinds of gated memory, e.g., long short-term memory units (LSTM) and gated recurrent units (GRU). The encoder turns the source sentence into
a sequence of semantic representations, or hidden states. During decoding, the attention mechanism aligns the state of the decoder to all the hidden states generated by the encoder, and decides which part of the input should be paid more attention.
By this information, the decoder can translate the semantic meaning of the input piece by piece.

A feature of this attention-based NMT model is that at each decoding step, the information of the
decoding history, e.g., the words that have been produced so far, is utilized to obtain a smooth translation. 
This is essentially a kind of language model. 
A potential problem here is that we only use the left context (decoding history), but ignores the right context (future words), although the right context 
could be valuable. This shortage can be partially alleviated by beam search, where the decision of the target word is 
delayed by a few steps, so the `future' information can be employed to impact the `past' decision. However, the 
potential of this beam search is rather limited, and we have found that most of the sequences in the buffer 
share the same prefix\cite{huang2008forest}. Thus a better solution is desired. 


In this paper, we propose a two-stage translation approach to tackle this problem. This approach is based on the idea of drafting-and-refinement, by which a draft translation is produced at the first stage, and at the second stage, the draft is refined by 
referring to the draft translation. Since the draft has given a rough
idea what the translation would be, the right context can be obtained and utilized to make the refinement. 
In our implementation, the first stage (drafting) uses a typical attention-based NMT system, and the second stage (refinement) uses
a double-attention NMT model that we will present shortly after. 

The remaining of the paper is structured as follows: the next section reviews some related work, and Section 3 briefly describes the attention-based NMT model. Section 4 introduces the double-attention NMT model, and Section 5 presents the experiments. Finally, 
the paper ends up with a conclusion.

\section{Related Work}

The idea of using the right context to aid translation has been used in several studies. Sutskever et al. found that his sequence-to-sequence  model achieved a promising improvement when reversing the source sentence ``$a,b,c$'' to ``$c,b,a$''\cite{Sutskever:14}. They argued that reversing the input may result in better memory usage during decoding, but it could also be possible that the right context is more informative when encoding the source input. The importance of the right context is also demonstrated by the fact that a significant improvement could be obtained when using a bi-directional RNN rather than using a uni-directional RNN\cite{schuster1997bidirectional}.

Recently, Novak et al. proposed an iterative translation approaches\cite{novak2016iterative}. Similar to our two-stage approach,
they got a draft translation using NMT, and then designed a `word correction' model that can correct the potential errors 
in the draft translation. 
The author raised the a similar argument that the right context is important to regularize the translation; the difference is that they focused on error
correction but we perform a complete new translation. Our approach may avoid the co-correction problem, i.e., correcting one 
word may impact the correctness of other words. 
 

The drafting-and-refinement idea was also used in other tasks. For example in automatic Chinese poetry composition\cite{yan2016poet},
a draft poem was firstly produced, and then the output was used as the input of the next iteration, to produce a new poem with better 
quality. The same approach was used in image generation\cite{gregor2015draw}, by which an image was drawn step-by-step, and the 
residual error was minimized at each step.




\section{Background: Attention-based NMT}

Our study is based on the attention-based NMT model\cite{2}, so we give a brief introduction for the sake of completeness. For
simplicity, our introduction is just the basic architecture presented in \cite{2}. Recent development of the attention-based NMT
using different architectures can be found in \cite{gehring2017convolutional,vaswani2017}.

This typical attention-based model is shown in Fig. 1, where the encoder and decoders are implemented as two RNNs. 
Put it in brief, a source sentence $X = (x_1, x_2, ..., x_{T_x})$ is encoded by the encoder RNN into a sequence of 
annotations $C = (h_1, h_2, ..., h_{T_x})$. Then the decoder RNN initiates a decoding process from a `start' symbol. 
At each decoding step $t$, the decoder computes the relevance between the decoder state $s_{t-1}$ and each 
annotation $h_i$, resulting in the attention weight $\alpha_{ti}$. The target word is generated by maximizing 
the conditional probability $p(y_t|y_{<t}, C, \alpha_{ti})$. 

\subsection{Encoder}

The encoder adopts the form of a bidirectional RNN (BiRNN)\cite{schuster1997bidirectional}, in which the hidden units can be either GRUs\cite{9} or  LSTMs\cite{8}. In this paper, we used the GRU units. This BiRNN decoder consists of a forward RNN $\overrightarrow{GRU}$ and a backward RNN $\overleftarrow{GRU}$. The forward RNN reads the source sentence from left to right and generates a sequence of forward 
annotations:
\[
\overrightarrow{C} = (\overrightarrow{h}_1, \overrightarrow{h}_2, ..., \overrightarrow{h}_{T_x}), 
\]

\noindent in which

\begin{equation}
\label{NMT_encoder}
\overrightarrow{h}_i = \overrightarrow{GRU}(x_i, \overrightarrow{h}_{i-1}).
\end{equation}

\noindent Similarly, the backward RNN reads the input sequence from right to left and generates a sequence of backward annotations:

\[
\overleftarrow{C} = (\overleftarrow{h}_1, \overleftarrow{h}_2, ..., \overleftarrow{h}_{T_x}).
\]

The final annotation $h_t$ is then obtained by a concatenation of $\overrightarrow{h}_t$ and $\overleftarrow{h}_t$, i.e., 

\[
h_t = [\overrightarrow{h}_{t}^{\top};\overleftarrow{h}_{t}^{\top}]^{\top}.
\]

\begin{figure}[t]
\begin{center}
\includegraphics[height=50mm]{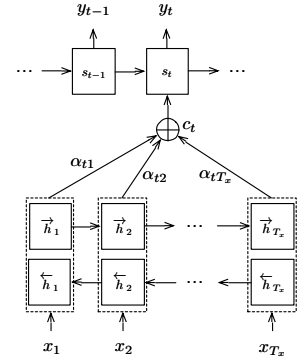}
\end{center}
\caption{The encoder-decoder NMT with attention.}
\vspace*{-3pt}
\end{figure}

\subsection{Attention}

When decoding the $t_{th}$ target word, the attention mechanism computes the attention weights:
 
\[
(\alpha_{t1}, \alpha_{t2}, ..., \alpha_{T_x})  = \sigma (e_{t1}, e_{t2}, ..., e_{T_x})
\]

\noindent where $\sigma(\cdot)$ is the softmax function, and 

\[
e_{ti} = f(s_{t-1}, h_i),
\]
where $s_{t-1}$ is the hidden state of the decoder at step $t$, and $f(\cdot)$ is the attention function 
that can be implemented by a neural network. The context vector $c_t$ is calculated as a weighted sum 
of annotations $C$, given by:

\begin{equation}
c_t = \sum_{i=1}^{T_x}\alpha_{ti}h_i.
\end{equation}

In this way, the decoder will pay attention to the annotations that are most relevant to the 
present decoding status, where the target-relatedness is represented by the attention weight $\alpha_{ti}$.

\subsection{Decoder}

As soon as we get the context vector $c_t$ from $C$ at decoding step $t$, the conditional probability of selecting a 
word $y_t$ is calculated as:

\begin{equation}
\label{g_function}
p(y_t|y_{<t}, C) = g(y_{t-1}, s_{t}, c_t),
\end{equation}
where $s_t$ is the the hidden state of the decoder at the $t_{th}$ step, and it is updated according to 
the previous hidden state $s_{t-1}$, the previous output $y_{t-1}$, and the context vector $c_t$:

\begin{equation}
\label{decoder_gru}
s_{t} = GRU(y_{t-1}, s_{t-1}, c_t).
\end{equation}

\subsection{Training}

All the parameters in the attention-based NMT model are optimized by maximizing the following conditional log-likelihood 
on the training dataset:

\begin{equation}
\mathcal{L}(\theta) = \sum_{n=1}^{N}\sum_{t=1}^{T^{(n)}_y}\log p(y^{(n)}_t|y^{(n)}_{<t}, X^{(n)}, \theta),
\end{equation}

\noindent where $X^{(n)}$ denotes the $n_{th}$ training sample, i.e., a bi-lingual sentence pair, and $\theta$ represents
the model parameters that we need to optimize. This optimization can be conducted by any numerical optimization approach,
but stochastic gradient descend (SGD) is the most often used. 

\section{Translation by Learning from Draft}

For the attention-based NMT, the posterior probability for the target word prediction is in the form $p(y_t|y_{<t}, X)$. 
Notice that it is conditioned on the entire source sentence $X$ and the decoding history $y_{<t}$, 
which is the left context. However, it does not involve any right context, although that information might be useful. 

One may argue that the backward information has been involved in the
annotations by the BiRNN encoding, therefore the right context information has been 
already taken into account. But this is not the case.
The right context we refer to is nothing to do with the semantic content that have been \emph{encoded}; instead, it is a 
regulation imposed by the \emph{target words} that would be \emph{decoded}.

We designed a two-state translation approach to solve the problem. By this approach, the source sentence $X$ is firstly translated into a \emph{draft} $\widetilde{Y} = (\widetilde{y}_1, \widetilde{y}_2, ..., \widetilde{y}_{T_{\widetilde{y}}})$ by an conventional attention-based NMT system, like the one in \cite{2}. Then a second-stage translation system will refine or `re-translate' this draft.
In this pipeline, the right context, although not very accurate, can be roughly obtained from the draft. This information will offer valuable regularization at the second-stage decoding, thus delivering a refined translation. In practice, we design a double-attention NMT model to utilize the right context information. This model accepts both the target draft $\widetilde{Y}$ and the original source sentence $X$, and pays attention to both the sequences during decoding. The main architecture is shown in Fig. 2.



\subsection{Encoder}

The double-attention model involves two encoders: the first encoder $GRU1$ serves to encode the source sentence $X$, and the second one $GRU2$ encodes the draft sentence $\widetilde{Y}$. Both encoders are BiRNNs and generate annotations. The formulations for the encoding are the same as (\ref{NMT_encoder}). At each encoding step $i$, the annotations $h_i$ and $\widetilde{h}_i$ are calculated as:

\begin{equation}
h_i = GRU1(x_i, h_{i-1}),
\end{equation}
\begin{equation}
\widetilde{h_i} = GRU2(\widetilde{y}_i, \widetilde{h}_{i-1}).
\end{equation}

\noindent Note that $GRU1$ and $GRU2$ both concatenate the forward and backward annotations. The two sequences of annotations are 
correspondingly written as  $C_1 = (h_1, h_2, ..., h_{T_x})$ and $C_2 = (\widetilde{h}_1, \widetilde{h}_2, ..., \widetilde{h}_{T_{\widetilde{y}}})$. 

\subsection{Attention}

\begin{figure}[t]
\begin{center}
\includegraphics[width=70mm]{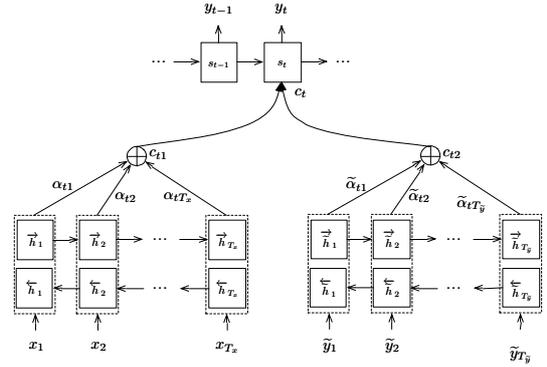}
\end{center}
\caption{The double-attention NMT model.}
\vspace*{-3pt}
\end{figure}

The double-attention model involves two attention mechanisms, one for the original source input and the other for the draft translation.
The final context vector is the concatenation of the context vectors on the two sequences:
\[
c_t = [c_{t1}^{\top};c_{t2}^{\top}]^{\top}
\]
where $t$ is the decoding step, $c_{t1}$ is the context vector produced by the attention mechanism on the original input, and the $c_{t2}$
is the context vector produced by the attention mechanism on the draft translation. These two context vectors are computed exactly as the 
attention mechanism of the conventional attention-based NMT model, as presented in the previous section.

\subsection{Decoder}

Using the concatenated context vector $c_t$, the decoder performs the translation as the conventional attention-based 
NMT: 

\begin{equation}
p(y_t|y_{<t}, C_1, C_2) = g(y_{t-1}, s_{t}, c_t),
\end{equation}
where $c_t = [c_{t1}^{\top};c_{t2}^{\top}]^{\top}$. The hidden state of the decoder is computed the same as (\ref{decoder_gru}), 
and the initial value of the hidden state is calculated as an average sum of the first backward annotations of the two input sequences:

\begin{equation}
s_0 = \frac{1}{2}(\overleftarrow{h_1} + \overleftarrow{\widetilde{h}}_1).
\end{equation}

\subsection{Training}

The training of the double-attention NMT model is similar to the conventional attention-based NMT model, though the log likelihood function now
depends on two input sequences $X$ and $\tilde{Y}$. This is written as follows:

\begin{equation}
\mathcal{L}(\theta) = \sum_{n=1}^{N}\sum_{t=1}^{T^{(n)}_y}\log p(y^{(n)}_t|y^{(n)}_{<t}, X^{(n)}, \widetilde{Y}^{(n)}, \theta).
\end{equation}

Note that to simplify the training, the architecture and the parameters of the first-stage NMT model can be inherited and re-used 
in the double-attention model. In our study, all the word embeddings (both on the source and target sides) are inherited from the 
first-stage NMT model and are fixed during the double-attention model training.

There are two reasons to keep these embeddings fixed. First of all, the embeddings have been well learned in the first stage, and re-using them in the second stage will significantly simplify the model training. The second and more important, the double-attention model consists of a large amount of  model parameters, which makes it prone to over-fitting, especially
when the training data is limited. We have observed the over-fitting problem on the small-scale task in our experiments, and re-using the word embeddings indeed reduced the over-fitting risk.


\section{Experiments}

\subsection{Datasets and evaluation metric}

The experiments were conducted on two Chinese-English translation tasks, one using the large-scale NIST dataset and the other using the small-scale IWSLT dataset. The NIST training data consisted of 1M sentence pairs, which involved 19M source tokens and 24M target tokens. 
We used the NIST 2005 test set as the development set and the NIST 2003 test set as the test set. The IWSLT training data consisted of 44K sentences sampled from the tourism and travel domain. The development set was composed of the ASR devset 1 and devset 2 from IWSLT 2005, and the test set was the IWSLT 2005 test set. As for the evaluation metric, we used the case-insensitive 4-gram NIST BLEU score\cite{papineni2002bleu}.

\subsection{Comparison systems}
We compared our two-stage system with two baseline systems: one is a conventional SMT system and the other is an attention-based NMT system (which is actually the first stage of our two-stage system). 

\subsubsection{Moses}
Moses\cite{koehn2007moses} is a widely-used SMT system and a state-of-the-art open-source toolkit. Although NMT has developed very quickly and outperforms SMT in some large-scale tasks, SMT is still a strong baseline for small-scale tasks. In our experiments, the following features were enabled for the SMT system: relative translation frequencies and lexical translation probabilities on both directions, distortion distance, language model and word penalty. For the language model, the KenLM toolkit\cite{heafield2011kenlm} was employed to build a 5-gram language model (with the Keneser-Ney smoothing) on the target side of the training data.

\subsubsection{Attention-based NMT}
We reproduced the attention-based NMT system proposed by Bahdanau et al.\cite{2}. The implementation was based on Tensorflow\footnote{https://www.tensorflow.org/}. We compared our implementation with a public implementation using Theano\footnote{https://github.com/lisa-groundhog/GroundHog/}, and got a comparable performance on the same data sets with the same parameter settings.

\subsection{Settings}

For a fair comparison, the configurations of the attention-based NMT system and the two-stage NMT system were intentionally set to be identical. The dimensionality of word embeddings, the number of hidden units and the vocabulary size were empirically set to 620, 1000, 30000 respectively for the large-scale task and were halved for the small-scale task. In the training process, we used the minibatch SGD algorithm together with the Adam algorithm\cite{kingma2014adam} to change the learning rate. The batch size was set to be 80. The initial learning rate was set to be 0.0001 for the large-scale task and 0.001 for the small-scale task. The decoding was implemented as a beam search, where the beam size was set to be 5.

\subsection{Results}

The BLEU results are given in Table I. It can be seen that our two-stage NMT system delivers notable 
performance improvement compared to the NMT baseline. On the large-scale task (NIST), the two-stage system
outperforms the NMT baseline by 0.9 BLEU points, and it also outperforms the SMT baseline by 1.1 points. 
On the small-scale task (IWSLT), the two-stage approach outperforms the NMT baseline by 2.4 BLEU points, though 
it is still worse than the SMT baseline (mainly because the SMT model is able to capture most details in the language pairs  while the NMT model tends to seize the generalities and treats rare details as noise, which is common when dataset is small). 
These results demonstrated that after the refinement with the double-attention
model, the quality of the translation has been clearly improved. 

\begin{table}[t]
\label{t1}
\begin{center}
\caption{BLEU scores on Chinese-English translation}
\begin{tabular}{l|c|c}
\hline
SYSTEM & NIST & IWSLT \\
\hline
\hline
Moses & 30.6 & \textbf{52.5} \\
\hline
Attention-based NMT & 30.83 & 43.83\\
\hline
Double-attention NMT & \textbf{31.71} & 46.32 \\
\hline
\end{tabular}
\end{center}
\end{table}

\section{Conclusions}

The attention-based NMT model performs the decoding from left to right, which can not fully utilize the right context. In this paper, we propose a two-stage translation approach that obtains a draft translation by a conventional NMT system, and then refines the translation by considering both the original input and the draft translation. By this way, the right context can be obtained from the draft and utilized to regularize the second-stage translation. Our experiments demonstrated that the two-stage approach indeed performs better than the conventional attention-based NMT system.
In the future work, we will investigate a better architecture to integrate the draft translation. Moreover, the memory usage of the double-attention model needs to be reduced.

\section*{Acknowledgment}
This work was supported by the National Natural Science Foundation of China under Grant No. 61371136 / 61633013
and the National Basic Research Program (973 Program) of China under Grant No. 2013CB329302.

\bibliography{reference}
\bibliographystyle{srt}

\end{document}